\documentclass[runningheads]{llncs}

\usepackage{eccv}

\usepackage{eccvabbrv}

\usepackage{graphicx}
\usepackage{booktabs}
\usepackage{amsmath}
\usepackage{amssymb}
\usepackage{mathtools}
\usepackage{url}
\usepackage{colortbl}    
\usepackage{xcolor}      
\usepackage{fontawesome5}
\usepackage{multirow}    
\newcommand{\Cow}[1][]{\includegraphics[width=10pt,trim={6cm 9cm 5cm 6cm},clip]{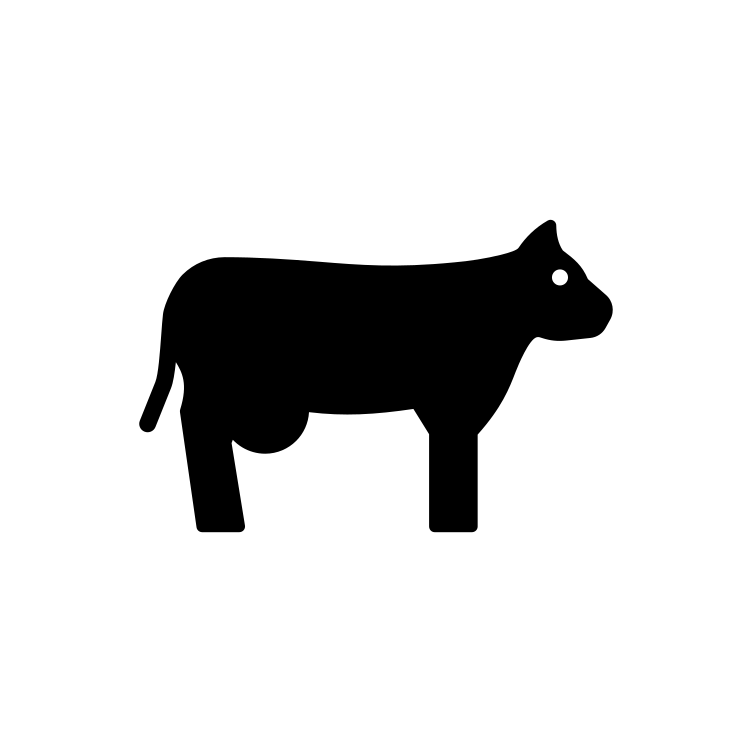}}
\newcommand{\Plant}[1][]{\includegraphics[width=10pt,trim={7cm 6cm 5cm 2cm},clip]{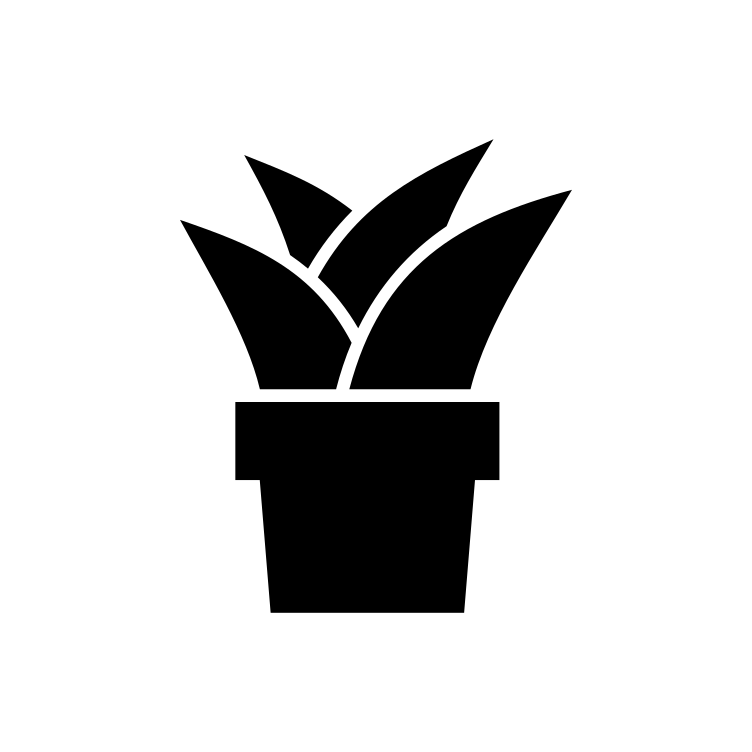}}
\newcommand{\Sheep}[1][]{\includegraphics[width=10pt,trim={6cm 7cm 5cm 6cm},clip]{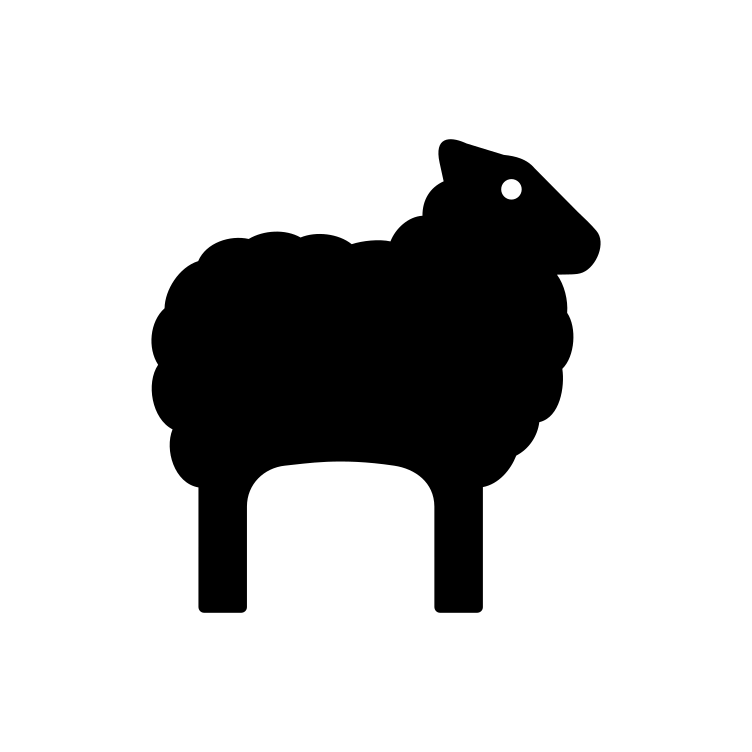}}

\newcommand{\best}[1]{\textbf{#1}}
\newcommand{\second}[1]{\underline{#1}}

\usepackage{colortbl}

\usepackage{bm}
\usepackage[normalem]{ulem}
\usepackage{float}

\usepackage[accsupp]{axessibility}  

\usepackage{hyperref}

\usepackage{orcidlink}

\begin{document}

\title{Towards Geometry-Grounded Dense Semantic Matching with VGGT Priors} 


\author{Songlin Yang\inst{1}\orcidlink{0000-0003-3403-376X} \and
Tianyi Wei\inst{2}\orcidlink{0000-0002-7976-8439} \and
Yushi Lan\inst{3}\orcidlink{0000-0003-4388-403X} \and
Zeqi Xiao\inst{4}\orcidlink{0009-0009-5694-8873} \and \\
Anyi Rao\inst{1}\orcidlink{0000-0003-1004-7753} \and
Xingang Pan\inst{4}\orcidlink{0000-0002-5825-9467}
}

\authorrunning{S. Yang et al.}

\institute{MMLab@HKUST, The Hong Kong University of Science and Technology, HK\and
Microsoft, USA \and
Visual Geometry Group, University of Oxford, UK \and
MMLab@NTU, Nanyang Technological University, Singapore}

\maketitle

\begin{abstract}
  Semantic matching aims to establish pixel-level correspondences between instances of the same category, and previous approaches based on 2D foundation models have achieved promising results on this task. However, they fail to achieve geometry-grounded dense semantic matching, which encompasses geometric awareness, manifold preservation, and cross-image invisibility reasoning. This new matching task simultaneously needs geometry-aware descriptors and holistic dense matching mechanisms, but existing approaches rarely provide both. To bridge this gap, we leverage VGGT, a 3D geometric foundation model that inherently provides strong geometry-grounded priors to support both capabilities in a unified framework. However, directly transferring faces two challenges: task heterogeneity, where VGGT originally matches cross-view images of the same instance, not cross-instance variations in shape or appearance; and scarce dense matching annotations for task adaptation. To address these challenges, we propose an approach that (i) retains VGGT’s intrinsic capabilities by reusing early feature stages, fine-tuning later ones, and adding a semantic head for bidirectional correspondences; and (ii) adapts VGGT for semantic matching under data scarcity through cycle-consistent training strategy, synthetic data augmentation, and progressive training recipe. Extensive experiments demonstrate that our approach achieves superior geometric awareness, manifold preservation, and matching reliability, outperforming previous baselines.
\end{abstract}

\section{Introduction}

\begin{figure}[t]
    \centering
    \includegraphics[width=\textwidth]{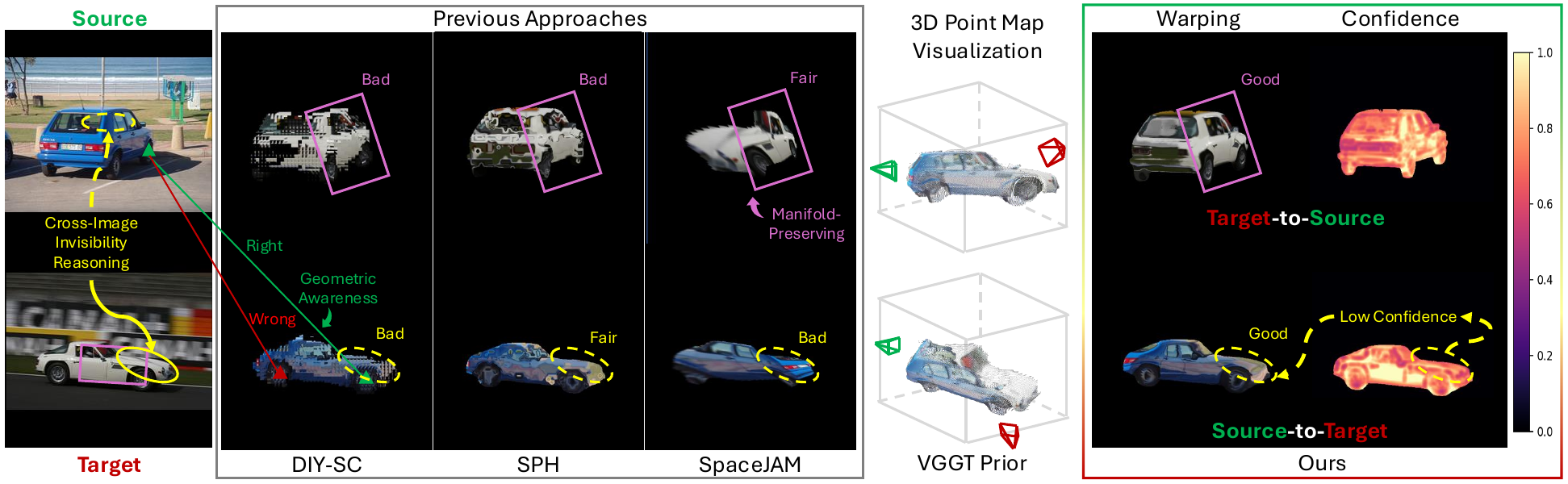}
    \caption{\textbf{Illustration of geometry-grounded dense semantic matching, limitations of previous approaches, and our improvements via leveraging VGGT priors.} Previous approaches using 3D prior with nearest-neighbor matching (\eg, DIY-SC~\cite{dunkel2025yourself}, SPH~\cite{mariotti2024improving}) or global alignment (\eg, SpaceJAM~\cite{barel2024spacejam}) each address only partially geometry-grounded requirements: \textcolor{green}{geometric awareness}, \textcolor{purple!60!white}{manifold preservation}, and \textcolor{yellow!90!black}{cross-image invisibility reasoning}. By visualizing the predicted 3D point maps, we find VGGT~\cite{wang2025vggt}, designed for geometry matching between views for one instance, can coarsely align same-category instances (more cases can be found in Fig.~\ref{fig: vggt}). VGGT also has inherent manifold-preserving matching capabilities. Built on VGGT, our approach not only achieves better geometry-aware and manifold-preserving matching, but also outputs prediction confidence for improving cross-image invisibility reasoning.}
    \label{fig:teaser}
    
\end{figure}

Semantic matching aims to establish pixel-level correspondences between semantically equivalent regions across two images with the same-category instances. Recent zero-shot works~\cite{zhang2023tale,dunkel2025yourself} show promising results. They leverage features from 2D foundation models, such as Stable Diffusion~\cite{rombach2022high,tang2023emergent} and DINO~\cite{oquab2023dinov2}, as pixel-level descriptors, and compute correspondences via nearest-neighbor matching (\ie, each pixel in the source image is matched to the pixel in the target image with the most similar feature).

However, practical downstream tasks (\eg, morphing~\cite{yang2025textured} and robotic affordance~\cite{lai2021functional}) have raised the more demanding requirements of geometry-grounded dense matching, as shown in Fig.~\ref{fig:teaser}, which include: (i) Geometric Awareness: disambiguating symmetric or repetitive structures such as front and rear wheels; (ii) Manifold Preservation: maintaining consistent relative positions among pixels when mapped from source to target (\eg, ensuring a source car window remains spatially coherent in the target image region, retaining its shape and internal layout rather than becoming distorted or fragmented); and (iii) Cross-Image Invisibility Reasoning: recognizing when a source pixel has no valid counterpart in the target image, typically due to occlusion or viewpoint-induced invisibility, which requires reasoning about 3D geometry rather than semantic similarity alone.

Previous approaches only partially achieve geometry-grounded matching. While some methods~\cite{zhang2024telling,mariotti2024improving,barel2024spacejam,mariotti2025jamais,fundel2024distillation,qian2025bridging} recognize the importance of geometric awareness and incorporate 3D priors to fine-tune 2D foundation model features, such as DIY-SC~\cite{dunkel2025yourself}, SPH~\cite{mariotti2024improving} in Fig.~\ref{fig:teaser}, they remain constrained by nearest-neighbor matching, failing to ensure manifold preservation and step towards cross-image invisibility reasoning (\eg, occlusions). Conversely, methods such as SpaceJAM~\cite{barel2024spacejam} in Fig.~\ref{fig:teaser}, that model global transformations to enforce manifold-preserving correspondences, often sacrifice the capacity for local non-rigid matching. These complementary shortcomings reveal that geometry-grounded semantic matching requires a unified approach that simultaneously provides geometry-aware pixel descriptors and holistic dense matching mechanisms.

Considering these geometry-grounded matching requirements, we observe that they naturally align with recent advances in 3D geometric foundation models such as DUSt3R~\cite{wang2024dust3r} and VGGT~\cite{wang2025vggt}. Although originally developed for 3D reconstruction from images, they provide strong geometry-aware features, and inherently learn image matching capabilities as a subtask of reconstruction~\cite{leroy2024mast3r}. As shown in Fig.~\ref{fig:teaser}, our preliminary evaluation of VGGT, a state-of-the-art reconstruction model, shows that it can coarsely align instances of the same category. Such capabilities make VGGT a highly promising basis for geometry-grounded dense semantic matching.

However, direct transferring faces two challenges: (i) Task Heterogeneity: VGGT was designed for geometric matching across-view images of the \textit{same instance}, which differs fundamentally from semantic matching across \textit{different instances} that exhibit variations in appearance and shape;
(ii) Data Scarcity: The lack of dense correspondence annotations further complicate its adaptation.

To address these challenges, we design an approach that combines \textit{capability retention} and \textit{task-specific adaptation}. (i) For retaining geometry-aware features and holistic matching, we reuse VGGT’s early feature stages, fine-tune its later ones, and append a semantic matching head that predicts dense bidirectional correspondence maps between source and target. (ii) For adaptation under limited annotations, we introduce a cycle-consistent training strategy that couples matching–reconstruction consistency with error–confidence correlation; curate a synthetic data pipeline generating diverse correspondences across categories and viewpoints; and adopt a progressive training recipe with aliasing artifact mitigation that gradually transfers dense correspondence ability from synthetic domains to real-world data. Extensive experiments and ablation studies demonstrate the superiority of our performance and the effectiveness of each key design.

Our contributions can be summarized as follows: (i) \textbf{\textit{Beyond ``Point Search'' (Paradigm Shift)}}: Traditional matching is bottlenecked by independent pixel-level nearest-neighbor search, which inherently fails at manifold preservation. We propose the first feed-forward framework unifying 3D-aware descriptors with a holistic dense matching rule. This enables demanding downstream tasks for geometry-grounded dense semantic matching that previous sparse matching simply cannot support. (ii) \textbf{\textit{Beyond ``Pseudo-3D''}}: Semantic matching is inherently a 3D problem involving symmetry and occlusion. Rather than adapting 2D features with pseudo-3D labels, we leverage a native 3D foundation model to bridge geometric and semantic variations, providing a more fundamental and innovative solution. (iii) \textbf{\textit{Beyond ``Ad-Hoc Engineering''}}: We resolve the long-standing scarcity of real-world dense correspondence labels via synthetic-to-real transfer. This may inspire the related tasks (\eg, 3D registration) in overcoming the same bottleneck: transitioning from sparse, instance-specific to dense, category-level correspondence.

\section{Related Works}
\textbf{Learning-Based Semantic Matching.}
Semantic matching is difficult due to appearance variations and scarce, ambiguous annotations~\cite{truong2021warp,zhang2025semantic}. Early methods relied on handcrafted feature descriptors~\cite{liu2010sift}, while the advent of deep learning facilitated the development of more effective feature extractors~\cite{yi2016lift,kim2017dctm,novotny2017anchornet,rocco2018end} and direct semantic correspondence detection networks~\cite{rocco2017convolutional,han2017scnet,kim2019semantic}. To address the limitations of scarce annotations, existing approaches have evolved three strategies: (i) leveraging weak supervision signals~\cite{lan2021discobox,chen2020show,zhang2023self,truong2022probabilistic,yang2024learning}, (ii) employing warping and cycle consistency constraints~\cite{zhou2016learning,truong2021warp,truong2022probabilistic,lan2022ddf_ijcv}, and (iii) augmenting pseudo-label supervision~\cite{kim2022semi,li2021probabilistic,huang2023weakly}.

\noindent
\textbf{2D-Foundation-Model-Based Semantic Matching.} Recent research has demonstrated the potential of leveraging features from 2D foundation models for zero-shot semantic correspondence prediction~\cite{amir2021deep,zhang2023tale,hedlin2023unsupervised,tang2023emergent,cheng2024zero, stracke2025cleandift}. Among these, DINO features~\cite{caron2021emerging,oquab2023dinov2} have been particularly effective~\cite{amir2021deep,zhang2023tale,zhang2024telling,suri2024lift,fundel2024distillation}. Diffusion model features~\cite{rombach2022high,stracke2025cleandift} offer complementary strengths~\cite{hedlin2023unsupervised,tang2023emergent,zhang2023tale,zhang2024telling,mariotti2024improving,li2024sd4match,fundel2024distillation,xue2025matcha}. However, simple nearest-neighbor search exhibits systematic limitations in disambiguating symmetric object parts~\cite{luo2023diffusion,zhang2024telling,mariotti2024improving,li2024sd4match,wimmer2024back,sommer2025common3d}. To address this, several studies have proposed appending adapter modules fine-tuned with pseudo or ground-truth labels~\cite{zhang2024telling,xue2025matcha,dunkel2025yourself}. Alternative approaches construct joint atlases for objects across multiple images~\cite{gupta2023asic,ofri2023neural}. \cite{zhang2024telling} through keypoint-specific information, but this cannot resolve all symmetries via simple image transformations (\eg, flipping) and requires keypoint-specific information that is generally unavailable. Approaches using 3D priors, such as DistillDIFT~\cite{fundel2024distillation}, SPH~\cite{mariotti2024improving}, and Jamais Vu~\cite{mariotti2025jamais}, show promise but remain limited for cross-instance and complex cases. 

\noindent
\textbf{3D Geometric Foundation Models.}
Recent works~\cite{zhang2025reviewfeedforward3dreconstruction} such as DUSt3R~\cite{wang2024dust3r} and VGGT~\cite{wang2025vggt} replace traditional multi-stage geometry pipelines with feed-forward transformers that directly predict 3D reconstruction signals (\eg, camera poses and 3D point maps). Building on these efforts, MASt3R~\cite{leroy2024mast3r} demonstrates strong image matching performance within the same scene. In contrast, we extend this line of research from purely geometric to semantic matching, aiming to establish dense correspondences across different object instances within the same category.

\section{Methodology}

We introduce our approach from: (i) Architecture Adaptation: We review VGGT and present our motivation and extension (Sec.~\ref{prediction}). (ii) Training: To adapt it to semantic matching with limited data, we propose a cycle-consistent training strategy (Sec.~\ref{optimization}), curate a synthetic data pipeline (Sec.~\ref{synthetic data}), and adopt a progressive training recipe (Sec.~\ref{progressive}) with aliasing artifact mitigation (Sec.~\ref{aliasing}).

\subsection{Semantic Correspondence Prediction with Geometry-Grounded Priors from VGGT}
\label{prediction}

\begin{figure}[th!]
    \centering
    \includegraphics[width=\textwidth]{ 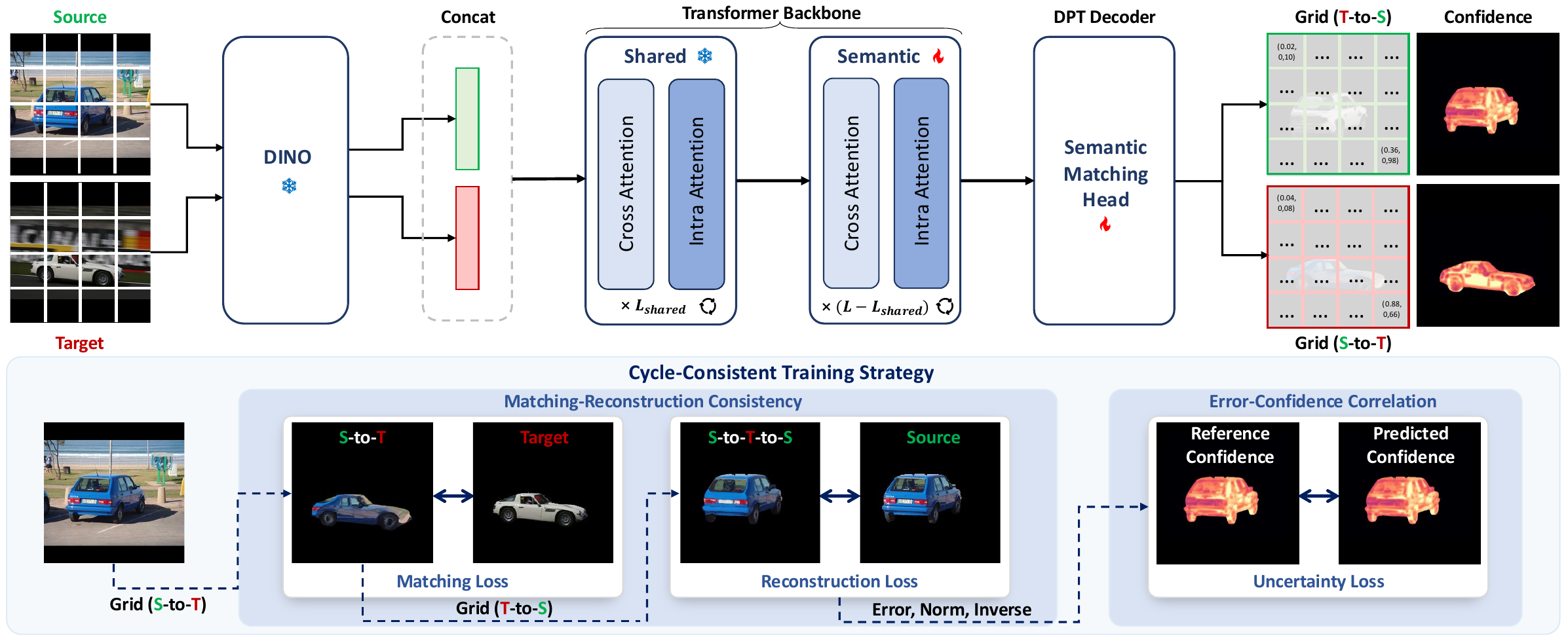}
    \caption{\textbf{Approach overview.} (i) Pipeline: The source and target images are first processed through a DINO-based feature extractor, followed by a VGGT-adapted backbone for feature refinement, and finally output pixel-wise correspondences in grid form via a semantic matching head. (ii) Cycle-Consistent Training Strategy: Matching–reconstruction consistency, ensuring that mapping an image to its counterpart and back reconstructs the original, and error–confidence correlation, aligning reconstruction error with predicted confidence to encourage reliable predictions.}
    \label{fig: framework}
\end{figure}

\textbf{VGGT Preliminary.} Given a set of $N$ RGB images $\{\bm{I}_i\}_{i=1}^N \in \!\mathbb{R}^{H\times W}$, VGGT applies a single feed-forward transformer backbone with $L$ blocks and DPT-based decoders~\cite{ranftl2021vision}, to predict camera parameters $\bm{\hat g}_i\!\in\!\mathbb{R}^{9}$ and pixel-level functional maps, such as depth $\hat {\bm{D}}_i\!\in\!\mathbb{R}^{H\times W}$ and 3D points $\hat {\bm{P}}_i\!\in\!\mathbb{R}^{3\times H\times W}$:
{\scriptsize
\begin{equation}
f:\ \{{\bm{I}}_i\}_{i=1}^{N}\ \longmapsto\ \big\{(\hat {\bm{g}}_i,\ \hat {\bm{D}}_i,\ \hat {\bm{P}}_i)\big\}_{i=1}^{N}.
\label{eq:vggt-map}
\end{equation}
}
Concretely, as shown in Fig.~\ref{fig: framework}, each image is first patchified by a DINO encoder into tokens, which are processed by alternating frame-wise inter attention (within each image) and global cross attention (across all images) in the transformer backbone. Tokens are then reshaped into low-resolution feature maps $\hat {\bm{F}} \in \mathbb{R}^{C\times H'\times W'}$ and upsampled with DPT decoders to produce $\hat {\bm{D}}_i$ and $\hat {\bm{P}}_i$.

\noindent
\textbf{Motivation of Leveraging and Extending VGGT.} Leveraging VGGT's priors and extending VGGT to semantic correspondence are well-motivated because: (i) Both cross-view geometric matching (same instance) and cross-instance semantic matching (same category) implicitly project images onto a canonical space, where instance-specific coordinates for geometry, category-level semantic landmarks for semantics. (ii) VGGT, trained for 3D reconstruction, extracts geometry-grounded representations from DINO’s semantic-geometric features while implicitly yielding a semantic branch. (iii) VGGT's manifold-preserving matching capability is essential to current semantic correspondence approaches.

\noindent
\textbf{Semantic Correspondence Prediction.}
Building on VGGT pipeline, we aim to predict dense semantic correspondences between a source image ${\bm{I}}_s$ and a target image ${\bm{I}}_t$. As shown in Fig.~\ref{fig: framework}, for the transformer backbone, we reuse the early $L_{shared}$ blocks (kept frozen) to extract geometry-grounded tokens, while fine-tuning the latter $(L-L_{shared})$ blocks to obtain semantic tokens. These are reshaped into feature maps $\hat {\bm{F}}_s,\ \hat {\bm{F}}_t \in \mathbb{R}^{C\times H'\times W'}$, which are fed into a new DPT-based semantic matching head $\phi_{\text{match}}$ to predict bidirectional sampling grids (\ie, grids denote the coordinates used to sample color values from one image when generating the warping image): $\hat {\bm{G}}_{s\!\to\!t}\in[-1,1]^{2 \times H\times W},\qquad
\hat {\bm{G}}_{t\!\to\!s}\in[-1,1]^{2\times H\times W}$,
where coordinates are normalized to $[-1,1]$, following the convention of $\mathrm{grid\_sample}$ function~\cite{paszke2019pytorch}. In addition, $\phi_{\text{match}}$ predicts pixel-wise confidence maps $\hat {\bm{C}}_{s},\hat {\bm{C}}_{t}\in[0,1]^{H\times W}$ by adding one dimension, which provides reliability estimation for correspondences.

\noindent
\textbf{Training Objectives.} The model is optimized through a progressive training recipe (Sec.~\ref{progressive}) with: (i) \textit{Supervised Loss ($L_2$ Loss)} from real data with sparse keypoints and synthetic data (Sec.~\ref{synthetic data}) with dense correspondence labels (\ie, grids); (ii) \textit{Cycle-Consistency Loss} (Sec.~\ref{optimization}) to refine matching and learn prediction uncertainty; and (iii) \textit{Smoothness Loss} to mitigate aliasing artifacts (Sec.~\ref{aliasing}).

\subsection{Cycle-Consistent Training Strategy}
\label{optimization}

Given scarce dense annotations and the need to handle cross-image invisibility, we
train the model to predict grids $\hat {\bm{G}}_{s\!\to\!t},\hat {\bm{G}}_{t\!\to\!s}$ and confidences $\hat {\bm{C}}_s,\hat {\bm{C}}_t$ with a cycle-consistent strategy, including matching-reconstruction consistency and error-confidence correlation.

\medskip\noindent\textbf{Matching-Reconstruction Consistency.}
Let $\mathcal{W}(\cdot,{\bm{G}})$ be $\mathrm{grid\_sample}$ function, we can obtain matching and reconstruction images:
{\scriptsize
\begin{align}
\hat{\bm{I}}_{s\!\to\!t} &= \mathcal{W}(\bm{I}_s,\hat {\bm{G}}_{s\!\to\!t}),\qquad
\hat{\bm{I}}_{t\!\to\!s} = \mathcal{W}(\bm{I}_t,\hat {\bm{G}}_{t\!\to\!s}), \\
\hat{\bm{I}}_{s\circlearrowleft} &= \mathcal{W}\!\big(\hat{\bm{I}}_{s\!\to\!t},\hat {\bm{G}}_{t\!\to\!s}\big),\qquad
\hat{\bm{I}}_{t\circlearrowleft} = \mathcal{W}\!\big(\hat{\bm{I}}_{t\!\to\!s},\hat {\bm{G}}_{s\!\to\!t}\big).
\end{align}}
We then use ${\bm{M}}_s,{\bm{M}}_t\in\{0,1\}^{H\times W}$ as object masks~\cite{ravi2024sam,lang_segment_anything}, and obtain the matching loss $\mathcal{L}_{\text{matching}}$ and reconstruction loss $\mathcal{L}_{\text{reconstruction}}$:
{\scriptsize
\begin{align}
\mathcal{L}_{\text{matching}}
=& {\bm{M}}_t \odot 
   \big\|  E(\bm{I}_t) - E(\hat{\bm{I}}_{s\!\to\!t}) \|_{2} \odot \hat{\bm{C}}_t 
+ {\bm{M}}_s \odot 
   \big\|  E(\bm{I}_s) - E(\hat{\bm{I}}_{t\!\to\!s}) \|_{2} \odot \hat{\bm{C}}_s, \\
\mathcal{L}_{\text{reconstruction}}
=& {\bm{M}}_s \odot 
   \big\| \bm{I}_s - \hat{\bm{I}}_{s\circlearrowleft} \big\|_{2} \odot \hat{\bm{C}}_s  
+ {\bm{M}}_t \odot 
   \big\| \bm{I}_t - \hat{\bm{I}}_{t\circlearrowleft} \big\|_{2} \odot \hat{\bm{C}}_t.
\end{align}}
where $E$ is the DINO feature extractor. We weight predictions using the confidence maps $\hat {\bm{C}}_s,\hat {\bm{C}}_t$ to account for prediction uncertainty during matching.

\medskip\noindent\textbf{Error-Confidence Correlation.}
We require the predicted confidence to be inversely correlated with the reconstruction error.
Define per-pixel errors
{\scriptsize
\begin{equation}
{\bm{e}}_s = \big\| \bm{I}_s - \hat{\bm{I}}_{s\circlearrowleft} \big\|_{2}, \qquad
{\bm{e}}_t = \big\| \bm{I}_t - \hat{\bm{I}}_{t\circlearrowleft} \big\|_{2}, 
\end{equation}
\begin{equation}
{\bm{e}}^{*}_s
= \frac{
    {\bm{e}}_s - \mathrm{min}({\bm{M}}_s\!\odot\!{\bm{e}}_s)
}{
    \mathrm{max}({\bm{M}}_s\!\odot\!{\bm{e}}_s)
    - \mathrm{min}({\bm{M}}_s\!\odot\!{\bm{e}}_s)
},\qquad
{\bm{e}}^{*}_t
= \frac{
    {\bm{e}}_t - \mathrm{min}({\bm{M}}_t\!\odot\!{\bm{e}}_t)
}{
    \mathrm{max}({\bm{M}}_t\!\odot\!{\bm{e}}_t)
    - \mathrm{min}({\bm{M}}_t\!\odot\!{\bm{e}}_t)
}.
\end{equation}}
and the reference confidences ${{\bm{C}}}_s = 1-{\bm{e^*}}_s$,
${{\bm{C}}}_t = 1-{\bm{e^*}}_t$.
We correlate the confidence and error as:
{\scriptsize
\begin{equation}
\mathcal{L}_{\text{uncertainty}}
= \big\| {\bm{M}}_s \odot ({\bm{C}}_s - \hat{{\bm{C}}}_s) \big\|_{1}
+  \big\| {\bm{M}}_t \odot ({\bm{C}}_t - \hat{{\bm{C}}}_t) \big\|_{1}
- \lambda_{\text{conf}} \cdot \left( \sum {\bm{M}}_s \odot \hat{{\bm{C}}}_s + \sum {\bm{M}}_t \odot \hat{{\bm{C}}}_t \right).
\end{equation}}

\subsection{Synthetic Data with Dense Annotations}
\label{synthetic data}

 We generate paired data as follows: (i) 3D Asset Generation: We select 18 categories from SPair-71k~\cite{min2019spair}, expand textual descriptions for each instance via ChatGPT~\cite{openai2025chatgpt}, and input these descriptions into Trellis~\cite{xiang2025structured}’s text-to-3D model to produce 3D assets. (ii) Rendering~\cite{ravi2020pytorch3d}: Each 3D asset is rendered from multiple viewpoints to generate RGB images and depth maps. The index of the corresponding 3D point for each pixel is recorded as an index map (serving as pixel-wise ground-truth labels). (iii) Multi-Condition Image Generation: For each rendered image, we extract a Canny~\cite{canny1986edge} map and combine it with the depth map and different textual descriptions to condition FLUX~\cite{flux2024}, synthesizing RGB images with diverse textures. (iv) Paired Data Construction: For view-aligned pairs, transformations (\eg, rotation, scaling) are applied to a canonical grid to generate warping grids (\ie, ${\bm{G}}_{s\!\to\!t}, {\bm{G}}_{t\!\to\!s}$), followed by steps (i)-(iii) to produce paired images. For view-unaligned pairs, warping grids are derived by matching index maps between different 3D assets, then steps (i)-(iii) are used to generate paired images. 
 
\subsection{Progressive Training Recipe}
\label{progressive}
During training, we optimize the model through a four-stage progressive strategy: (i) Synthetic Data Pretraining (Dense Supervision): Train exclusively on synthetic data with dense ground-truth annotations, employing only the $L_2$ loss and smoothness loss (Sec.~\ref{aliasing}). This stage aims to equip the model with VGGT's manifold-preserving mapping capabilities that generalize across different instances. (ii) Real Data Adaptation (Sparse Keypoint Supervision): Introduce real data with sparse ground-truth keypoints by adding a keypoint $L_2$ loss to the existing losses. This facilitates the transfer of dense mapping capabilities from synthetic to real-world domains. (iii) Matching Refinement (Matching and Reconstruction): Further optimize by incorporating the matching loss $\mathcal{L}_{\text{matching}}$ and reconstruction loss $\mathcal{L}_{\text{reconstruction}}$ to enhance matching precision. (iv) Uncertainty Learning (Confidence Prediction): Finally, integrate the uncertainty loss $\mathcal{L}_{\text{uncertainty}}$ to learn error-dependent confidence, enabling the model to predict prediction reliability. 
\subsection{Aliasing Artifacts and Mitigation}
\label{aliasing}
The matching head predicts continuous coordinates for discrete pixels, causing aliasing artifacts, visible as checkerboard patterns (Fig.~\ref{fig: ablation}, 6th column), due to the inherent ambiguity (\ie, slight coordinate shifts in either direction remain plausible). We mitigate this with a smoothness loss that enforces spatial coherence by reshaping the grid into a vector and penalizing differences between adjacent coordinates.

\section{Experimental Settings}

\subsection{Implementation Details}

 We use a VGGT-based Transformer with $L{=}24$ blocks, where the first 4 blocks are fixed, and the remaining 20 blocks are duplicated from VGGT to initialize for a new semantic branch. Additionally, a DPT decoder is added as the semantic matching head, extracting features from blocks [4, 11, 17, 23]. The model is trained for 5 days using a single A6000 GPU. The evaluation is performed on SPair-71k~\cite{min2019spair}, AP-10k~\cite{yu2021ap} (intra-species (I.S.), cross-species (C.S.), and cross-family (C.F.)) and our synthetic dataset.

\subsection{Evaluation Metrics}


\noindent
\textbf{(i) Matching Accuracy (From Sparse to Dense)}. For \textit{sparse matching}, we adhere to standard protocols~\cite{gupta2023asic,huang2022learning,zhang2024telling,xue2025matcha,mariotti2024improving,min2019spair} and evaluate performance using the \textbf{(PCK)}. PCK@$\alpha$ is defined as the ratio of correctly predicted matched keypoints that lie within a radius of $R=\alpha \cdot \max(H,W)$ around their ground-truth positions, where $H, W$ refer to the height and width of the object's bounding box, respectively. For \textit{dense matching}, we evaluate the accuracy on the synthetic test set by measuring the Mean Squared Error (MSE) between the ground-truth images and the source images warped via predicted correspondences (\textit{i.e.}, \textbf{Synthetic Dense}). Given the lack of dense annotations in real-world data, we utilize \textbf{DINO Accuracy} and \textbf{SD Accuracy} (defined as the MSE between the feature maps of the target and the warped source) as direct proxies to measure.

\noindent
\textbf{(ii) Geometric Awareness}. To assess geometric awareness, we specifically compile results on symmetric keypoints (marked with an asterisk * in Tab.~\ref{tab:per-image-pck}) on SPair-71k, which directly tests the model's ability to disambiguate repetitive or left-right symmetric structures. 

\noindent
\textbf{(iii) Manifold Preservation}. To holistically evaluate manifold preservation on real data, we compute the \textbf{FID} between target and warped source images. Assuming ideal cycle consistency ($\hat{I}_{s\circlearrowleft} \approx I_s$) implies geometric consistency and local invertibility, we further measure reconstruction fidelity via \textbf{PSNR} (signal-level), \textbf{SSIM} (structural), and \textbf{LPIPS} (perceptual). Together, these metrics quantify the model’s manifold-preserving capabilities.


\noindent
\textbf{(iv) Cross-Image Invisibility Reasoning}. We perform two specialized quantitative analyses on the synthetic dataset: (iv-a) AUC with Different Invisibilities (\textbf{AUC - Horizontal Rotation}): We calculate the Area Under the Curve (AUC) for occlusion detection across varying horizontal rotation angles, where increasing rotation serves as a direct proxy for greater cross-image invisibility due to occlusions. (iv-b) Confidence Evaluation (\textbf{MSE - Confidence}): We utilize our predicted confidence map as a universal reference to bin the matching errors of all evaluated methods into corresponding confidence intervals, verifying whether the confidence scores effectively reflect matching reliability and identify non-corresponding pixels.

\begin{figure}[!t]
    \centering
    \includegraphics[width=\textwidth]{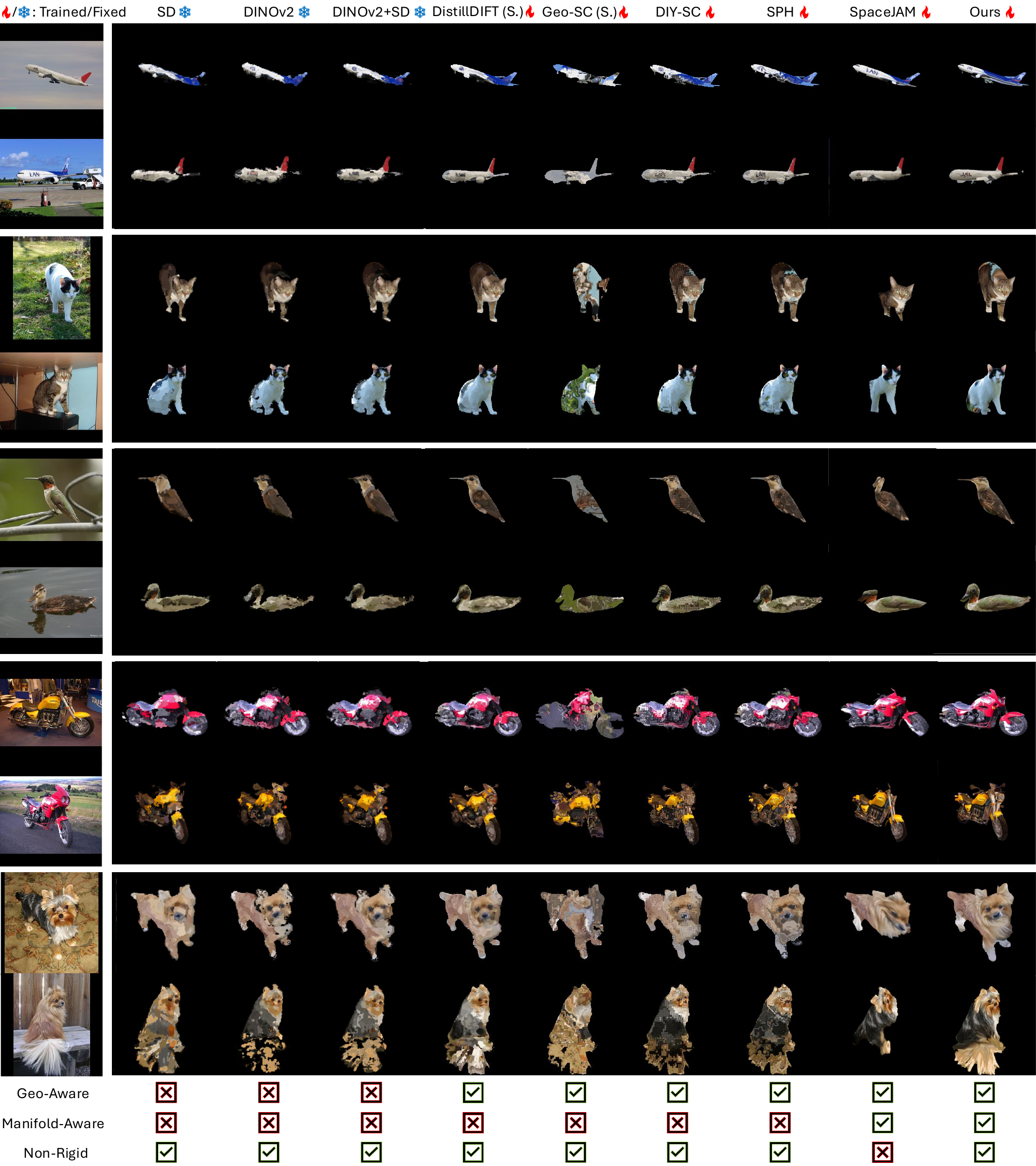}
    \vspace{-0.2cm}
    \caption{\textbf{Qualitative results of dense semantic matching compared with previous approaches.}}
    \label{fig: baseline}
    \vspace{-0.2cm}
\end{figure}

\begin{figure}[th]
    \centering
    \includegraphics[width=\textwidth]{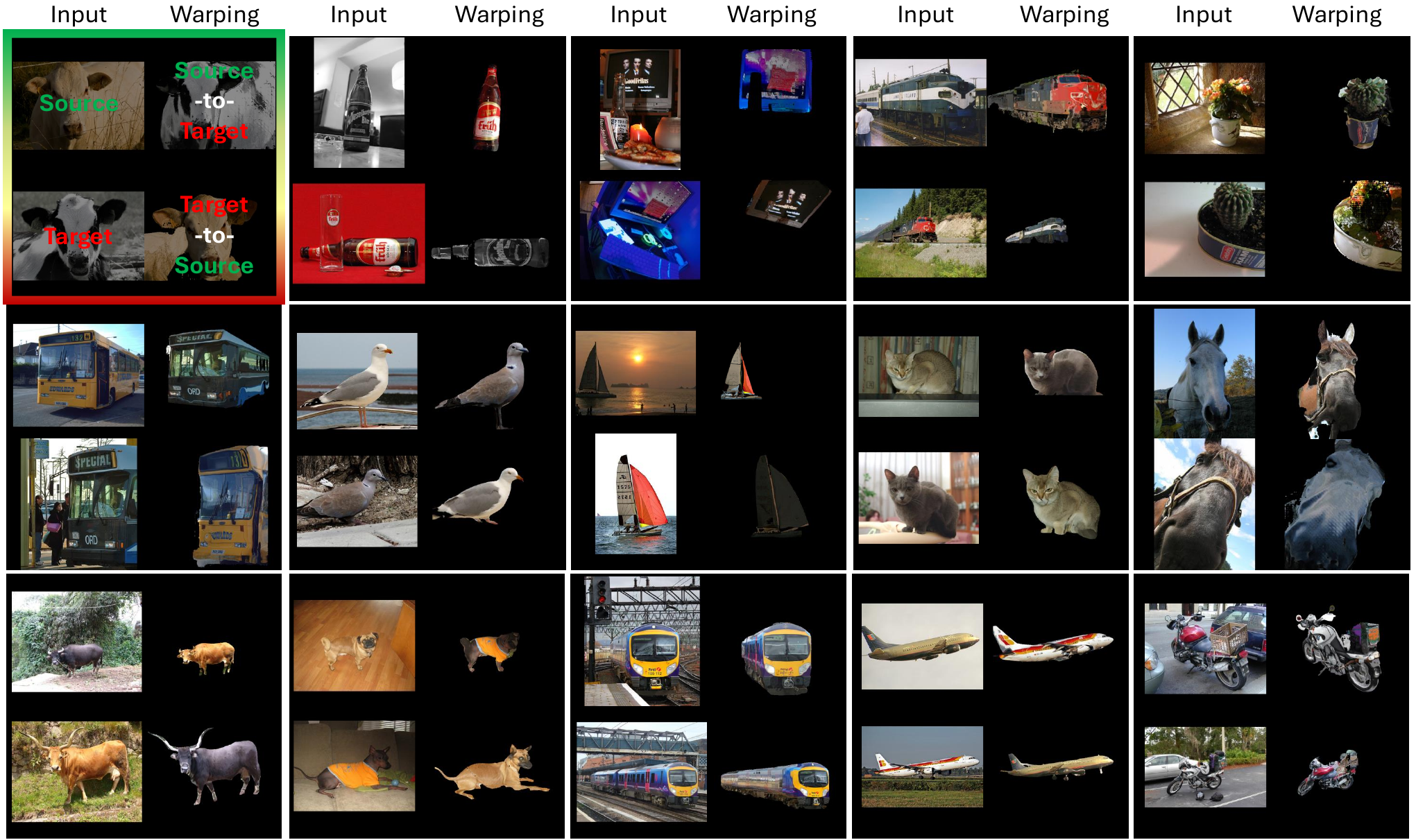}
    \caption{\textbf{More cases of dense semantic matching by our approach.} These results highlight our superior generalization across diverse object categories, viewpoints, and non-rigid deformations.}
    \label{fig: more ours}
\end{figure}

\subsection{Baseline Selection: Comprehensiveness and Fairness}

We select baselines from: (i) Training Requirements: SD~\cite{tang2023emergent}, DINO~\cite{oquab2023dinov2}, and SD+DINO~\cite{zhang2023tale} are training-free, whereas Geo-SC~\cite{zhang2024telling} (Flip = No sparse keypoint tuning; S. = Sparse keypoint tuning applied), DIY-SC~\cite{dunkel2025yourself}, SPH~\cite{mariotti2024improving}, ASIC~\cite{gupta2023asic}, and SpaceJAM\cite{barel2024spacejam} involve training. Notably, DistillDIFT~\cite{fundel2024distillation} (U.S. = No sparse keypoint tuning; S. = Sparse keypoint tuning applied) utilizes 3D synthetic data for training. Geo-SC, DIY-SC, SPH, ASIC, and SpaceJAM are trained with the same dataset as ours, while DistillDIFT uses more 3D synthetic data. (ii) Canonical Space Assumptions: SPH, ASIC, and SpaceJAM all adopt canonical space hypotheses. SPH explicitly requires same-category instances to share a spherical canonical space, while ASIC and SpaceJAM learn warp and transformations projected onto a canonical space, respectively.

\section{Matching Evaluation}

\subsection{Dense Matching}

Unlike sparse keypoint matching, which focuses only on a few salient points, dense semantic matching requires pixel-wise correspondences across the entire image, demanding geometric awareness, manifold preservation, and robustness to local non-rigidity. We present qualitative comparisons with baseline approaches in Fig.~\ref{fig: baseline}, and analyze across four critical dimensions: (i) Training Requirements: While fine-tuned approaches consistently outperform zero-shot approaches, Geo-SC (S.) demonstrates limited effectiveness for dense matching despite improvements in keypoint matching. This stems from its keypoint-centric training objective that neglects dense correspondence. (ii) Geometric Awareness: Approaches incorporating geometric regularization, learning canonical spaces, or data augmentation exhibit superior performance in distinguishing symmetric and repetitive regions, a critical challenge for semantic matching. (iii) Manifold Preservation: With the exception of SpaceJAM and our approach, existing approaches fail to maintain underlying manifold structures during matching. This limitation compromises their utility for downstream tasks (\eg, affordance learning) that require precise preservation of original surface geometries. (iv) Local Non-Rigidity: While SpaceJAM's global transformation learning preserves manifolds, its inability to handle non-rigid deformations restricts performance on objects with complex topologies. Only our approach satisfies all the requirements for dense semantic correspondence while addressing these fundamental limitations. Additional quantitative evaluations appear in Tab.~\ref{tab:per-image-pck}, with extended qualitative results presented in Fig.~\ref{fig: more ours}.

\begin{table}[t]

  \centering
    \caption{\textbf{Quantitative results on SPair-71k~\cite{min2019spair} and AP-10k~\cite{yu2021ap}.} 0.05* indicates additional aggregation on symmetric keypoints (\eg, front/rear wheels).}
  \setlength{\tabcolsep}{5pt} 
    \resizebox{\linewidth}{!}{
      \begin{tabular}{lccccccccccccccc}
        \toprule[1.5pt]
        \multirow{2}{*}{Models}&&   \multicolumn{4}{c}{SPair-71k (PCK~$\uparrow$)}& \multicolumn{3}{c}{AP-10k (PCK@0.1~$\uparrow$)}&Synthetic~$\downarrow$& DINO~$\downarrow$&SD~$\downarrow$&\multirow{2}{*}{FID~$\downarrow$}&\multirow{2}{*}{PSNR~$\uparrow$}&\multirow{2}{*}{SSIM~$\uparrow$}&\multirow{2}{*}{LPIPS~$\downarrow$}\\
            \cmidrule(lrr){3-6}\cmidrule(llr){7-9}
             & &  0.1 & 0.05 & 0.01 &0.05* &  I.S & C.S. & C.F.&Dense&Accuracy&Accuracy&\\
        \midrule
        SD + DINO & \cite{zhang2023tale} &  59.9 & 44.7 & 7.9& 40.3 & 62.9 & 59.3 & 48.3&0.20&0.35&0.28&28.73& 17.88 & 0.59 & 0.249\\
        DistillDIFT (U.S.) & \cite{fundel2024distillation} &  60.8& 45.4&8.0& 44.7&  65.8&64.2&56.1&0.16&0.21&0.18&32.04&18.08&0.62&0.242\\
        Geo-SC (Flip) & \cite{zhang2024telling} &  65.4& 49.1 &9.9& 51.0&  68.7& 64.6& 52.7&0.14&0.23&0.16&29.45&18.02&0.63&0.245\\
        \midrule
        DistillDIFT (S.) & \cite{fundel2024distillation} &  65.3 & 49.9& 8.9& 49.8 & 66.9&64.7&58.0&0.15&0.17&0.16&\second{18.98}& 18.95 & 0.63 & 0.221\\
        Geo-SC (S.) & \cite{zhang2024telling} & \second{82.9} & \second{72.6}  &\second{21.6} & \second{73.5}&  70.1& 68.3& \second{58.4}& 0.26& 0.51&0.28&128.70& 15.72 & 0.49 & 0.318\\
        DIY-SC & \cite{dunkel2025yourself} &    71.6 & 54.2 &10.1& 53.8&\second{70.6} &\second{69.1} & 57.8&0.11&0.15&0.14&25.46& 17.22 & 0.56 & 0.268 \\
        SPH & \cite{mariotti2024improving} &  64.4& 48.2&8.4& 49.3&  65.4&63.1&51.0&0.10&0.17&0.15&23.89& 16.51 & 0.53 & 0.289\\
        
        \midrule
        ASIC& \cite{gupta2023asic} &  36.9& 34.2& 5.8& 32.3&  32.5&29.1&23.0&0.09&0.11&0.10&29.21&14.99&0.50&0.287\\
        SpaceJAM & \cite{barel2024spacejam} &  44.5& 34.6& 6.7& 33.8&  42.7&39.9&35.2&\second{0.08}&\second{0.10}&\second{0.07}&23.06& \second{21.13} & \second{0.71} & \second{0.178}\\
        \rowcolor[HTML]{E6F2FF}
        Ours & &  \best{84.2} & \best{76.3} & \best{25.9} & \best{77.8} & \best{73.2}  & \best{71.4} & \best{60.9} & \best{0.07}&\best{0.06}& \best{0.06}&\best{3.58}& \best{23.84} & \best{0.78} & \best{0.142}\\
        \bottomrule[1.5pt]
      \end{tabular}
    }
    \label{tab:per-image-pck}
\end{table}

\subsection{Sparse Keypoint Matching}

Sparse keypoint matching evaluates the ability to match semantically salient regions with geometric awareness. Although previous approaches~\cite{fundel2024distillation,mariotti2024improving} attempted 3D synthetic data augmentation, they still struggle with geometry-ambiguous regions. In contrast, our approach surpasses previous approaches on this task, as shown in Tab.~\ref{tab:per-image-pck}.

\noindent
\textbf{``Keypoint Overfitting'' Trap.} We observe an interesting phenomenon that solely overfitting sparse salient points neglects global manifold preservation. Despite high PCK, Geo-SC (S.) suffers in dense matching quality; in contrast, our method ensures superior fidelity. The 3D prior's robustness is confirmed on AP-10k, where both DIY-SC and ours outperform Geo-SC (S.), despite the latter's higher sparse PCK on SPair-71k. This illustrates the importance of our proposed holistic semantic matching paradigm and the introduction of 3D priors.

\begin{figure}[t]
    \centering
    \includegraphics[width=\textwidth]{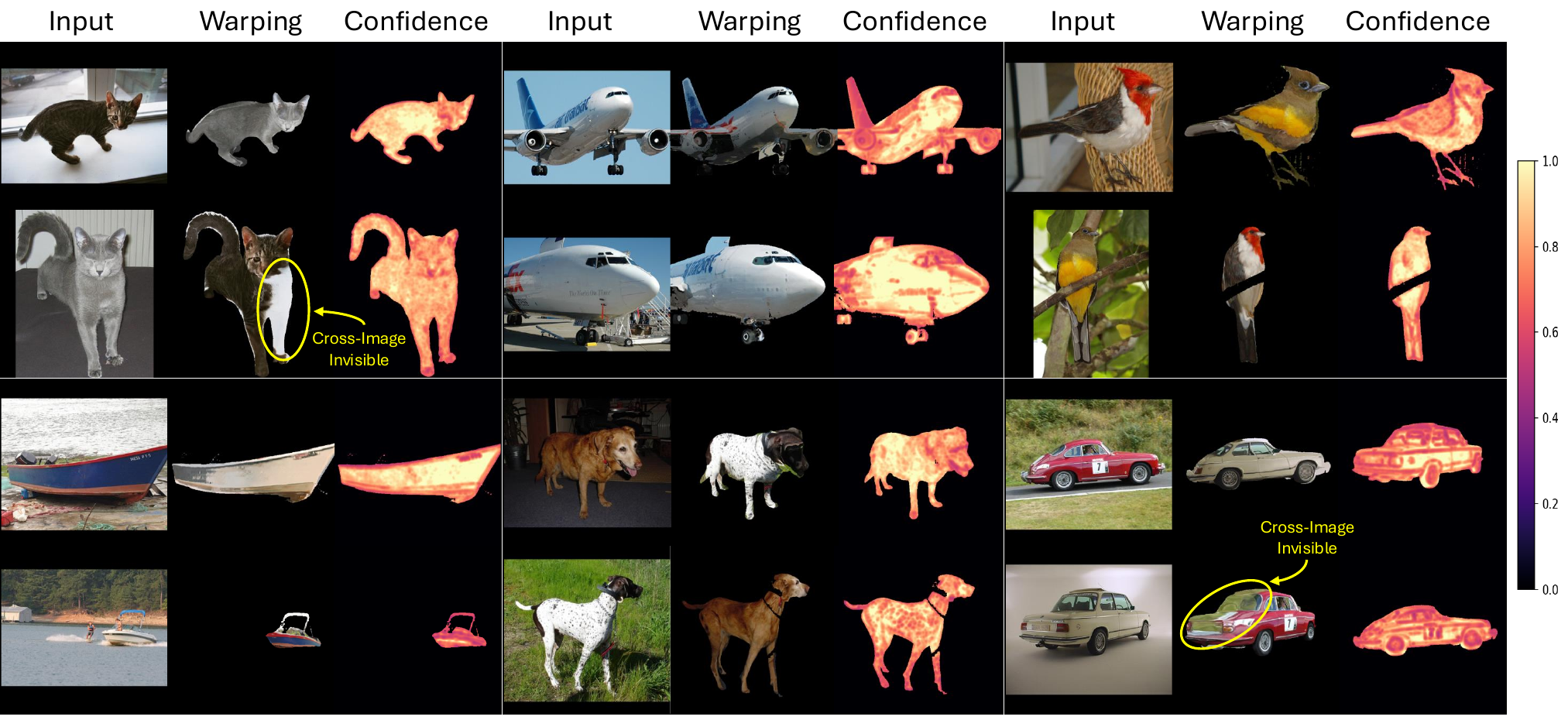}
    \caption{\textbf{More qualitative results of our approach on prediction reliability.} Confidence map prediction is promising to addresse the overlooked issue of cross-image invisibility, enhancing optimization via uncertainty calibration and offering reliability cues for downstream tasks.}
    \label{fig: confidence}
\end{figure}

\begin{table}[t]
    \caption{\textbf{Quantitative evaluation for invisibility \& confidence on synthetic dataset with dense ground truth.} Horizontal rotation angles denote the object's viewpoint deviation in the target image relative to the source. Confidence intervals represent predicted matching reliability, with higher values indicating greater certainty.}
    \centering
    \setlength{\tabcolsep}{15pt} 
    \resizebox{\linewidth}{!}{
      \begin{tabular}{lcccccccccc}
        \toprule[1.5pt]
        \multirow{2}{*}{Models}&&   \multicolumn{5}{c}{AUC - Horizontal Rotation $\uparrow$}& \multicolumn{4}{c}{MSE - Confidence (\%) $\downarrow$}\\
            \cmidrule(lrr){3-7}\cmidrule(llll){8-11}
            &&  $5^\circ$ & $30^\circ$ & $45^\circ$ &$90^\circ$ & Mean & 90-100 & 80-90 & 60-80& 0-60\\
        \midrule
        SD + DINO &\cite{zhang2023tale}& 0.55  & 0.54 & 0.48 &0.37  & 0.52&0.140 & 0.162 & 0.189 & 0.230 \\
        DistillDIFT (U.S.) &\cite{fundel2024distillation}& 0.62 & 0.61 & 0.57 & 0.54& 0.60& 0.124 & 0.128 & 0.131 & 0.183 \\
        Geo-SC (Flip) &\cite{zhang2024telling}& 0.64  & 0.62 & 0.53 & 0.50& 0.61& 0.123 & 0.126 & 0.130 & 0.164 \\
        \midrule
        DistillDIFT (S.) &\cite{fundel2024distillation}& 0.72  & 0.70 & 0.69 & 0.66& 0.69& 0.122 & 0.129 & 0.134 & 0.161 \\
        Geo-SC (S.) &\cite{zhang2024telling}& 0.59  & 0.52 & 0.48 &0.42 & 0.51& 0.182 & 0.189 & 0.195 & 0.296 \\
        DIY-SC &\cite{dunkel2025yourself}&  0.75 & 0.71 & 0.70 & 0.68& 0.71& 0.097 & 0.100 & 0.108 & 0.125\\
        SPH &\cite{mariotti2024improving}& 0.70  &0.65  & 0.62 &0.59 &0.64 & 0.081 & 0.092 & 0.101 & 0.126 \\
        \midrule
        ASIC &\cite{gupta2023asic}& 0.82 & 0.80  & \second{0.75} & \second{0.72}& 0.78& 0.082 & 0.089 & 0.095 & 0.112  \\
        SpaceJAM &\cite{barel2024spacejam}& \second{0.85} & \second{0.81} & 0.73 & 0.70& \second{0.81}& \second{0.078} & \second{0.083} & \second{0.089} & \second{0.108} \\
        \rowcolor[HTML]{E6F2FF}
        Ours & & \best{0.95} & \best{0.93} & \best{0.91} & \best{0.89} & \best{0.92} & \best{0.064} & \best{0.069} & \best{0.070} & \best{0.091}\\
        \bottomrule[1.5pt]
      \end{tabular}
      }
    \label{tab:invisible}

\end{table}

\subsection{Matching Reliability: Invisibility \& Confidence}

 We identify cross-image invisibility, where source pixels lack valid counterparts in the target image, as a critical yet overlooked factor that degrades matching reliability. Motivated by this, as shown in Fig.~\ref{fig: confidence}, we propose confidence map prediction, which not only introduces uncertainty calibration during matching learning to enhance optimization but also provides a reliability reference for downstream applications to selectively adopt high-confidence correspondences. To further evaluate invisibility quantitatively, we use AUC with different invisibilities. As shown in Tab.~\ref{tab:invisible}, ours consistently maintains the highest AUC across all angles. We then use our confidence map as a universal reference for all methods, and calculate MSE for the corresponding confidence interval based on the confidence value. All methods show an increase in MSE as confidence decreases. This consistent correlation proves that our confidence map truly identifies occlusions and non-corresponding pixels.

\section{Ablation Study}

\begin{figure}[t]
    \centering
    \includegraphics[width=\textwidth]{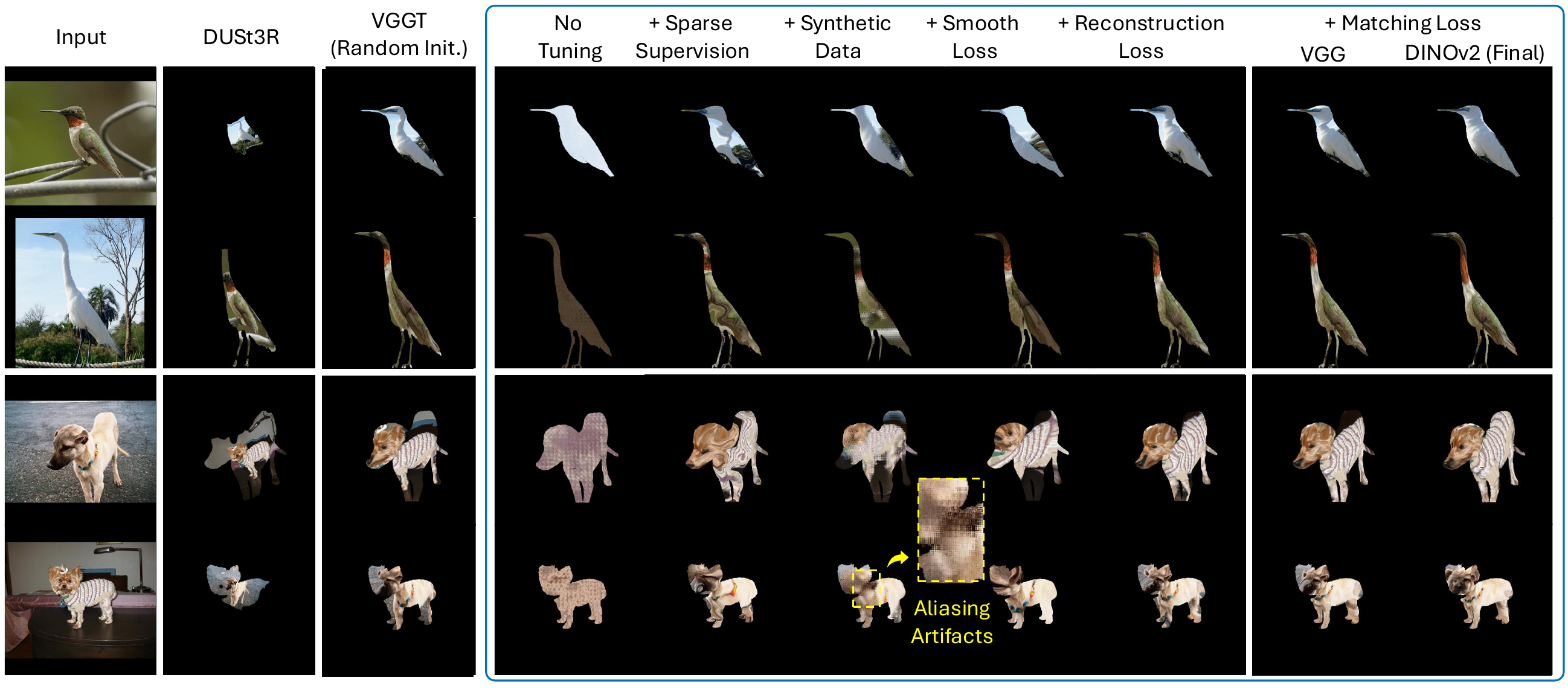}
    \caption{\textbf{Qualitative ablation results of the proposed approach.} Ablation study on two key aspects: (i) 3D reconstruction pretraining, showing that DUSt3R and randomly initialized VGGT are inferior to our approach in semantic matching accuracy and training efficiency; and (ii) training strategy design, where stage-wise incorporation of synthetic data, sparse supervision, smoothness, reconstruction, and matching losses highlights the effectiveness of our final training recipe.}
    \label{fig: ablation}
\end{figure}

\noindent
\textbf{Effectiveness of Each Key Design.} As shown in Fig.~\ref{fig: ablation} and Tab.~\ref{tab:ablation}, we assess the effectiveness of our proposed approach through two key aspects: (i) 3D Reconstruction Pretraining Priors: We compare with DUSt3R~\cite{wang2024dust3r} and randomly initialized VGGT. VGGT extends DUSt3R by using DINO as input and richer training data. DUSt3R shows manifold-preserving ability but lacks semantic richness due to the absence of 2D foundation model features (\eg, DINO), while random VGGT achieves basic matching but requires far more refinement than our approach. (ii) Training Strategy Analysis: Our analysis reveals that directly using VGGT's geometric correspondences without tuning fails for semantic matching. The introduction of sparse supervision enables basic matching but compromises manifold preservation. Incorporating synthetic data restores manifold structure yet introduces aliasing artifacts. While smoothness loss mitigates these artifacts, it reduces matching accuracy. Further addition of reconstruction loss improves performance but remains suboptimal. We compare VGG loss~\cite{johnson2016perceptual} and DINO loss, ultimately selecting the latter as the optimal matching loss due to its superior fine-grained semantic matching accuracy. 

\begin{table*}[t]
  \centering
  \setlength{\tabcolsep}{9pt} 
  \begin{minipage}[t]{0.55\columnwidth}
    \caption{
      \textbf{Ablation of the proposed training recipe.} We evaluate the contribution of each component step-by-step.
    }
    
    \centering
    \resizebox{\linewidth}{!}{
      \begin{tabular}{lcc}
        \toprule[1.5pt]
        Settings & SPair-71k (PCK@0.1) $\uparrow$ & Synthetic Dense $\downarrow$ \\
        \midrule
        DUSt3R & 58.7 & 0.15 \\
        VGGT (Random Init.) & 69.2 & 0.13 \\
        No Tuning & 9.0 & 0.49 \\
        + Sparse Supervision & 79.4 & 0.18 \\
        + Synthetic Data & 79.3 & 0.11 \\
        + Smoothness Loss & 79.1 & 0.10 \\
        + Reconstruction Loss & 80.5 & 0.10 \\
        + Matching Loss (VGG) & 81.8 & 0.09 \\
        \rowcolor[HTML]{E6F2FF}
        + Matching Loss (DINO) & \textbf{84.2} & \textbf{0.07} \\
        \bottomrule[1.5pt]
      \end{tabular}
    }
    \label{tab:ablation}
  \end{minipage}
  \hfill
  \begin{minipage}[t]{0.44\columnwidth}
    \caption{
      \textbf{Ablation of VGGT adaptation for shared blocks and DPT layers.}}
    
    \centering
    \resizebox{\linewidth}{!}{
      \begin{tabular}{cccc}
        \toprule[1.5pt]
        \multicolumn{2}{c}{Models} & SPair-71k & Synthetic $\downarrow$ \\
        Shared & DPT Layers & PCK@0.1 $\uparrow$ & Dense \\
        \midrule
        18 & [4,11,17,23] & 59.3 & 0.22 \\
        12 & [4,11,17,23] & 65.0 & 0.14 \\
        6 & [4,11,17,23] & 77.2 & 0.09 \\
        \rowcolor[HTML]{E6F2FF}
        4 & [4,11,17,23] & \textbf{84.2} & \textbf{0.07} \\
        4 & [3,10,16,22] & 80.1 & 0.09 \\
        4 & [2,9,15,21] & 78.9 & 0.11 \\
        4 & [1,8,14,20] & 76.4 & 0.10 \\
        \bottomrule[1.5pt]
      \end{tabular}
    }
    \label{tab:layer}
  \end{minipage}
\end{table*}

\begin{figure}[t]
    \centering
    \includegraphics[width=\textwidth]{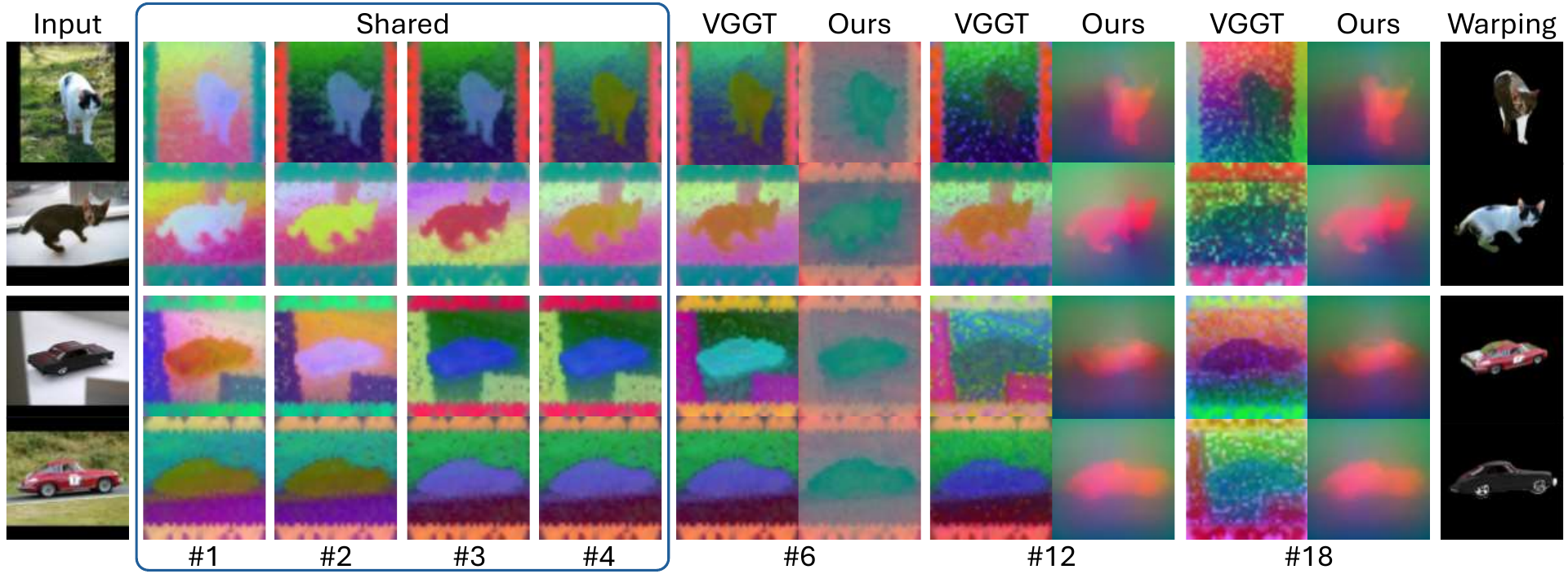}
    \caption{\textbf{Feature visualization of original VGGT backbone and our adapted semantic branch.} Unlike the original VGGT branch, which yields noisy cross-instance activations, our fine-tuned semantic branch produces coherent, semantically aligned correspondences.}
    \label{fig: Block}
\end{figure}

\begin{figure}[t]
    \centering
    \includegraphics[width=\textwidth]{ 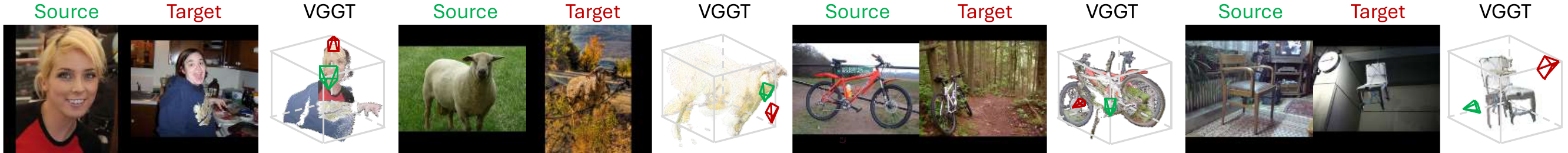}
    \caption{\textbf{More VGGT-predicted cases.}}
    \label{fig: vggt}
\end{figure}

\noindent
\textbf{Backbone Block Selection and Feature Visualization.} We propose an effective architectural adaptation that augments VGGT with semantic matching while preserving its extensibility for broader downstream tasks. We conduct in-depth analyses about this adaptation on two key aspects: (i) Backbone Adaptation Analysis: Testing the blocks of transformer backbone, we ultimately adopt the first 4 blocks as shared components and retain features from blocks [4, 11, 17, 23] for DPT input. Quantitative validations are provided in Tab.~\ref{tab:layer}. (ii) Feature Visualization: We employ Principal Component Analysis (PCA) to project features from different blocks into three RGB channels for visualization. Visual comparisons reveal that VGGT’s original branch struggles with cross-instance image pairs, exhibiting weak feature correlations and noisy activations. However, our fine-tuned semantic branch demonstrates clear matching coherence and smooth feature distribution, as shown in Fig.~\ref{fig: Block}. 

\noindent
\textbf{More VGGT Cases: Coarse Cross-Instance Alignment.} To further validate our motivation regarding VGGT's capability for coarse alignment, we supplement additional examples in Fig.~\ref{fig: vggt}.

\section{Conclusion}


In this work, we have pioneered a framework for feed-forward dense semantic matching, fundamentally shifting the task from independent point-wise retrieval to holistic structural inference. By moving beyond the ``Nearest Neighbor'', our approach ensures global manifold consistency and structural coherence, enabling the model to handle complex non-rigid deformations that traditional sparse matching paradigms fail to resolve. This transition from ``searching for similarity'' to ``inferring geometric flow'' provides a more promising foundation for dense semantic correspondence. We introduce a novel paradigm by repurposing 3D geometric foundation models for semantic correspondence. Our approach suggests that the native coupling of 3D structural priors and semantic descriptors offers a promising pathway to tackle symmetry and invisibility. By moving beyond conventional 2D-centric tuning, we leverage intrinsic 3D priors to effectively bridge the gap between geometry and semantics.

\noindent
\textbf{Limitations, Scalability and Future Work.} (i) More Views: Although we propose a novel VGGT-based semantic matching approach, multi-view consistent semantic matching across scenes remains an open challenge. (ii) More Foundation Models: For compatibility with VGGT and fair baseline comparisons, we limit feature extraction to DINOv2. Future work could integrate additional features from other foundation models (\eg, DINOv3 and Stable Diffusion). (iii) More Categories: To overcome the persistent scarcity of dense annotations, we have established a principled data-centric roadmap that synergizes synthetic-to-real transfer with cycle-consistent self-supervision. This strategy could be further explored on datasets with more diverse object categories. (iv) More Extreme: Beyond its current auxiliary role, our uncertainty modeling offers a path toward robust matching under extreme viewpoints and occlusions.

\section{Acknowledgment}

This work was supported by NTU SUG-NAP and S-Lab. This work was also partially supported by a grant from the NSFC/RGC Collaborative Research Scheme Project No. CRS\_HKUST605/25.



%
%
\bibliographystyle{splncs04}
\bibliography{main}
\end{document}